%% file: icml2023.tex
\documentclass{article}

\usepackage{microtype}
\usepackage{graphicx}
\usepackage{subfigure}
\usepackage{comment}
\usepackage{booktabs} 
\usepackage{amsfonts}
\usepackage{adjustbox}
\usepackage{amsmath}
\usepackage{hyperref}
\usepackage{threeparttable} 

\usepackage[accepted]{icml2023}

\usepackage[capitalize,noabbrev]{cleveref}

\newcommand\blfootnote[1]{%
  \begingroup
  \renewcommand\thefootnote{}\footnote{#1}%
  \addtocounter{footnote}{-1}%
  \endgroup
}
\newcommand{\maysam}[1]{\textcolor{black}{#1}}
\newcommand{\mb}[1]{\textcolor{black}{#1}}

\newcommand{\maxk}[1]{\textcolor{black}{#1}}

\newcommand{\MK}[1]{\textcolor{black}{#1}}
\newcommand{\MKFinal}[1]{\textcolor{black}{#1}}

\usepackage[textsize=tiny]{todonotes}

\icmltitlerunning{TIDE: Time Derivative Diffusion for Deep Learning on Graphs}

\begin{document}

\twocolumn[
\icmltitle{TIDE: Time Derivative Diffusion for Deep Learning on Graphs}



\icmlsetsymbol{equal}{*}

\begin{icmlauthorlist}
\icmlauthor{Maysam Behmanesh}{equal,to}
\icmlauthor{Maximilian Krahn}{equal,to,goo}
\icmlauthor{Maks Ovsjanikov}{to}

\end{icmlauthorlist}

\icmlaffiliation{to}{LIX, École polytechnique, IP Paris, France}
\icmlaffiliation{goo}{Aalto University, Finland}

\icmlcorrespondingauthor{Maysam Behmanesh}{maysam.behmanesh@lix.polytechnique.fr}
\icmlcorrespondingauthor{Maximilian Krahn}{maximilian.krahn@icloud.com}

\icmlkeywords{Machine Learning, ICML}

\vskip 0.3in
]

\printAffiliationsAndNotice{\icmlEqualContribution}




\begin{abstract}
A prominent paradigm for graph neural networks is based on the message-passing framework.
In this framework, information communication is realized only between neighboring nodes.
The challenge of approaches that use this paradigm is to ensure efficient and accurate \textit{long-distance communication} between nodes, as deep convolutional networks are prone to oversmoothing.
In this paper, we present a novel method based on time derivative graph diffusion (TIDE) to overcome these structural limitations of the message-passing framework.
Our approach allows for optimizing the spatial extent of diffusion across various tasks and network channels, thus enabling medium and long-distance communication efficiently. Furthermore, we show that our architecture design also enables local message-passing and thus inherits from the 
\MKFinal{capabilities}
of local message-passing approaches.
We show that on both widely used graph benchmarks and synthetic mesh and graph datasets, the proposed framework outperforms state-of-the-art methods by a significant margin.$^+$
\end{abstract}

\input{sec-introduction}
\input{sec-related}
\input{sec-motivation.tex}

\input{sec-method}
\input{sec-experiments}
\input{sec-conclusion}

\bibliography{icml2023}
\bibliographystyle{icml2023}

\input{sec-appendix.tex}

\end{document}

%% file: sec-introduction.tex
\section{Introduction}

Designing efficient and scalable architectures for learning on graphs is a central problem in machine learning with applications in a broad range of disciplines, including data mining \cite{li2019graph,zhang2019attributed}, recommendation systems \cite{zhang2019attributed}, text classification \cite{yao2019graph}, image analysis and matching \cite{sarlin2020superglue} and even molecular property prediction \cite{wieder2020compact} among myriad others.

A very wide variety of graph neural network (GNN) approaches have been proposed over the past several years (see, e.g., \cite{zhou2020graph,wu2020comprehensive} for recent surveys), ranging from spectral methods, spatial or convolutional designs, recurrent graph neural networks, or graph auto-encoders as well as many other hybrid techniques. A particularly prominent and widely-used category of approaches is given by the convolutional graph neural networks, and especially those based on message-passing, following the design introduced in \cite{Kipf:2017tc} and extended significantly in many follow-up works, e.g., \cite{li2018adaptive,zhuang2018dual,chamberlain2021grand,thorpe2021grand++}.

The key strengths of convolutional graph neural networks, as introduced in \cite{Kipf:2017tc}, include their simplicity and computational efficiency, their ability to be composed with other neural networks as well as their ability to generalize across different graphs (i.e., learning weights that could be applied on unseen graphs). As a result, the original GCN approach \cite{Kipf:2017tc} is still highly effective and is widely used in many applications.

Nevertheless, a prominent limitation of message-passing approaches, such as GCN and related methods is \textit{oversmoothing}, which implies that such networks tend to be difficult to train beyond a small number of layers \cite{oono2019graph}. 
Furthermore, since typical message-passing operators only ensure communication between nodes within a $1$-hop neighborhood,  this means that message-passing approaches can hinder \textit{long-distance information propagation}, which can limit their utility in scenarios, where such long-range communication is important. 

In this work, we demonstrate that a simple modification to the standard GCN design can be used to enable information propagation across possibly distant nodes within a \MK{shallow, one layer} graph neural network, \MK{which does not have the oversmoothing issues of traditional deep GNNs.}
%
%
Our design inherits most of the advantages of standard message-passing methods, including computational efficiency, their domain independence, and their ability to generalize across different graphs. 
\blfootnote{\hspace*{-\footnotesep}$^+$Our implementation is available at \url{https://github.com/maysambehmanesh/TIDE}}

Key to our approach is our use of \textit{learnable time diffusion} which allows information propagation on the graph while being able to optimize the communication extent in a task-dependent manner. Our method is inspired by recent approaches in surface learning \cite{sharp2022diffusionnet} that have introduced the notion of \textit{learnable time diffusion} as a way to replace convolution for information sharing when learning on surfaces. However, unlike the method presented in \cite{sharp2022diffusionnet}, which explicitly aimed at robustness to changes in connectivity and used standard heat diffusion, we base our approach on learnable \textit{time-derivative} diffusion. As we demonstrate in this work, this allows us to retain the advantages of the message-passing framework, and ensure both local \MK{($1-2$ hop neighborhoods)} and possibly global \MK{($n$-hop neighborhoods,  up to the whole graph)} information communication in a single an efficiently learnable architecture.
To summarize, our key contributions include:
\begin{enumerate}
	\item We introduce \textit{time-derivative diffusion} as an effective mechanism for information propagation within graph neural networks.
	\item We propose an architecture based on time-derivative diffusion, which enables both local and global information propagation, in a differentiable manner, while generalizing the standard message-passing framework.
	\item With this mechanism at hand, we develop a simple and scalable architecture that  outperforms strong baselines on several benchmarks. 
\end{enumerate}

Our method is particularly useful on either sparsely labeled graphs or in scenarios where longer dependencies are important to capture global structures on graphs. Our method is easy to train and does not require a significant computational overhead in either learning time or memory footprint. Finally, while we focus on the GCN and derived architectures, we believe that the use of learnable time-derivative diffusion can be a broadly useful tool in graph neural networks and can, in the future, be combined with other architecture designs.
%


%% file: sec-related.tex
\section{Related Work}
\paragraph{Graph neural networks}

%
%
%
The key goal of Graph Neural Networks (GNNs) is to compute the representation of all unlabeled nodes or edges.
To achieve this, many network designs employ a message-passing approach to exchange information of each node or edge with its neighbors until reaching the equilibrium state \cite{Scarselli4700287}. 
Based on the approach, GNNs can perform convolution using two types of models: spatial or spectral.
Spatial GNNs perform message passing directly by considering neighborhood structure in the graphs \cite{wu2020comprehensive}. 
\maxk{Below we list several well-known spatial GNNs:} GraphSage uses an aggregation function to represent each node by aggregate results of its neighborhood \cite{graphsage2017},
Message Passing neural network (MPNN) runs $K$-step message-passing iterations to let information propagate further \cite{gilmer2017neural}, and Graph Attention Networks (GAT) use an attention mechanism to combine different contributions of neighboring nodes \cite{velivckovic2017graph}. One of the most prominent models for message passing is the Graph Convolutional Network (GCN) \cite{Kipf:2017tc}. 
GCN simplifies ChebNets architecture \cite{Chebnet2016} by using filters operating on the $1$-hop neighborhoods of the graph. Within GCN, the features from the neighbors get passed to the node by convolution layers.

\par
%
Spectral GNNs perform convolution by transforming node representations into the spectral domain using the graph Fourier transform \cite{wu2020comprehensive}. GNN-ARMA \cite{bianchi2021graph} uses an autoregressive moving average (ARMA) filter to capture the global graph structures.
%
%
\maysam{To implement an efficient convolution on the graph, \cite{xu2018graph,behmane2021} consider the convolution via wavelet transform instead of Fourier transform by taking the graph wavelet as a set of bases of spectral GNN. 
}

\MK{A common problem on graph neural networks is their oversmoothing behavior, which hinders their expressive power when increasing the number of layers, \cite{li2018deeper}. 
Multiple approaches have tried to solve this issue, e.g., by using different co-training strategies \cite{li2018deeper}, changing the architecture by adding different kinds of residual connections \cite{li2019deepgcns}, a PageRank-based propagation schema \cite{klicpera2018predict} or a separate analysis of propagation and representation of the features \cite{liu2020towards}.
}
%
%

\paragraph{Learned diffusion / learned ODEs}
A stepping stone in deep neural networks was interpreting these as neural ordinary differential equations (ODE) \cite{chen2018neural}.
This idea has been further expanded in numerous works such as \cite{dupont2019augmented, finlay2020train,li2020scalable, liu2019neural}.
For graph neural networks, this interpretation has been applied in various works, for instance in \cite{avelar2019discrete, poli2019graph, xhonneux2020continuous}.
Additionally, ODEs have not only been used for graph neural networks but also for shape learning.
In geometric learning, Sharp et al. introduced a diffusion on shapes with the learnable time parameter $t$ \cite{sharp2022diffusionnet}. 

\paragraph{Graph diffusion process}
\mb{GRAND \cite{chamberlain2021grand} is a prominent graph diffusion work that interprets graph convolution networks as a solution to the heat diffusion equation.
This work forms the basis for models such as GRAND++ \cite{thorpe2021grand++}, which uses an additional source term, and BLEND \cite{chamberlain2021beltrami}, which uses additional Beltrami features. However, in these models, the diffusion time is treated as a fixed hyperparameter and is not learnable. This limits the flexibility and adaptability of such models to different graph structures.}

\paragraph{Long range dependency} \maxk{Commonly-used message-passing GNNs do not accommodate long-distance communication, which limits their expressivity to a small neighborhood around the node. In \cite{alon2020bottleneck} a novel explanation for training is introduced to prevent over-squashing in GNNs from long-range patterns in the data. In \cite{abu2019mixhop}, these relationships are learned by repeatedly mixing feature representations of neighbors at various distances.
}

%% file: sec-motivation.tex
\section{Background, Motivation \& Overview}
\label{sec:motivation}

Our work builds upon the successful paradigm in graph neural networks based on message passing. Methods within this framework, pioneered by the GCN architecture \cite{Kipf:2017tc} and its variants \cite{bianchi2021graph,Ming2020} are based on two key components. First, an information message passing operator $L$ is assembled, typically using a normalized Laplacian or adjacency matrix. Second, this operator is used, jointly with a non-linearity function $\sigma$, to construct a single layer of the graph neural network with learnable linear weights $W$. Such layers can then be stacked into a deep multi-layer network by iterating the action of a single layer and using separate learnable weights at each level.

Given a graph consisting of $n$ nodes, \MKFinal{let $U \in \mathbb{R}^{n \times m}$ be a matrix of $m$ scalar fields $u \in \mathbb{R}^n$ representing, for example, some feature values at the nodes}. The most basic variant of this approach can be summarized via the following formula:
\begin{align}
	\label{eq:graph_nn}
\mathcal{N}_{\mathcal{W}} (U) =  \mathcal{L}_k \circ \mathcal{L}_{k-1} ... \circ \mathcal{L}_1 \circ \mathcal{L}_0 (U).
\end{align}
Here $U$ is some set of input features, $\mathcal{L}_k$ is the $k^{\text{th}}$ layer of the neural network, and  $\mathcal{W}$ denotes the set of all learnable weights, which are composed of weights associated with every layer. In particular, a typical layer $\mathcal{L}_k$ has the form:
\begin{align}
	\label{eq:layer_gcn}
	\mathcal{L}_{k} (U) =  \sigma\left(L U W^k\right).
\end{align}
where $\sigma$ is some non-linearity, $W^k$ is a matrix of learnable weights at layer $k$ and $L$ is a message passing operator. 
For example, $L$ can be the standard graph Laplacian matrix or the normalized adjacency matrix with self-loops as used in \cite{Kipf:2017tc}.

While simple and efficient, this approach has several key drawbacks. Perhaps the most prominent limitation is the well-known \emph{oversmoothing} effect of the basic graph neural network architecture. This effect implies that networks of the type described in Eq.~\eqref{eq:graph_nn} tend to saturate very quickly, even for small to moderate $k$. In other words, it is difficult to build networks that are both easy to train and have a significant depth.

Since the most commonly-used message-passing operators such as the graph Laplacian or its normalized variants only enable information propagation within the $1$-hop neighborhood of each node, this means that standard graph neural networks do not easily enable \emph{long-distance information communication}, which can limit their utility in practice. Unfortunately, simple strategies such as expanding the receptive field size of the message passing operator also have limited success.

\paragraph{Motivation}
Our work aims to address the issue raised above and is inspired by recent approaches that exploit properties of the \textit{diffusion} process to enable communication on both graphs \cite{chamberlain2021grand,thorpe2021grand++} and more general domains \cite{sharp2022diffusionnet}. Specifically, in  \cite{chamberlain2021grand} the authors showed that information propagation within a graph neural network can be formulated from the perspective of anisotropic diffusion, which furthermore encompasses and generalizes the standard message-passing formulation. However, in that work, the diffusion time was still used as a fixed hyperparameter (set to 1).

On the other hand, with the computer graphics community \cite{sharp2022diffusionnet} it has recently been shown that diffusion with \emph{a learnable time} parameter can be used to enable information propagation on geometric domains. Moreover, this process can adapt the receptive field size of different channels from local to global depending on the task. In particular, rather than using message passing, the key idea in \cite{sharp2022diffusionnet} is to ensure information propagation by using the diffusion equation:
\begin{align}
	\label{eq:diffusion}
	\frac{\partial u}{\partial t} = - \Delta u,
\end{align}
where $\Delta$ is a positive semi-definite Laplacian operator. The solution to the diffusion equation is given by the heat operator:
\begin{align}
	\label{eq:heat_op}
u_0 \rightarrow u_t, \text{ where } u_t = H_t(u_0)
\end{align}
Thus, the idea advocated in \cite{sharp2022diffusionnet} is to use Eq.~\eqref{eq:heat_op} to build a layer, where each signal $u$ is propagated for some task-dependent, \textit{learnable time} $t$. Notably, the heat operator $H_t$ has a closed form expression and is given by the operator (matrix) exponential $H_t = \exp(-t \Delta)$ and can moreover be approximated using the Laplacian spectral basis for efficient computation.

On the other hand, the architecture design of the approach in \cite{sharp2022diffusionnet} 9s focused on ensuring robustness to significant discretization changes, especially on triangle meshes. This means that the connectivity structure of the underlying graph is not explicitly relied upon. In contrast, within graph neural network applications, the connectivity structure is often crucial and is a major source of useful information, which, in part, also explains the success of message-passing approaches.

Our main goal thus is to combine the local accuracy and sensitivity to the graph structure of message-passing approaches, with the ability to learn the receptive field size and ensure global information propagation without oversmoothing, enabled by diffusion with learnable time parameters. 

\paragraph{Overview}
To achieve these goals, we propose a novel model, which combines learnable diffusion with message passing in a single principled framework while being efficient and achieving accurate results on a wide range of benchmarks.

Our method is based on the diffusion equation, Eq. \eqref{eq:diffusion}, similar to the approach in  \cite{sharp2022diffusionnet}. However, rather than using diffusion itself to ensure information propagation on the graph, we propose to use \textit{time-derivative diffusion}. That is, rather than using Eq~\eqref{eq:heat_op} to propagate information on a graph within a graph neural network, we propose to use the following time derivative formulation instead:
\begin{align}
	\label{eq:deriv_heat_op}
	u_0 \rightarrow -\frac{\partial{u_t}}{{\partial t}}, \text{ where } u_t = H_t(u_0).
\end{align}
Our key idea, therefore, is to use Eq.~\eqref{eq:deriv_heat_op} to enable information communication in a graph within a graph neural network. Despite a relatively simple change, as we demonstrate below, this formulation has several distinguishing characteristics. First, conceptually, time-derivative diffusion is closely linked to wavelets, since, for example, it is well known that the derivative in time of the heat kernel, which is simply a Gaussian in Euclidean space, corresponds exactly to the Mexican hat wavelet \cite{hou2012continuous,kirgo2021wavelet}.
Moreover, and more importantly, as we demonstrate below, unlike standard diffusion, a time-derivative-based formulation allows us to retain the power of local message-passing approaches while, at the same time, enabling long-distance communication without oversmoothing.

%% file: sec-method.tex
\section{Method}
\subsection{Time Derivative Diffusion}
\label{subsec:tide_method}
As mentioned above, in the continuous setting, the diffusion process is described as the solution of the heat equation, Eq.~\eqref{eq:diffusion}: $\frac{d}{d t} u_t= L u_t$,
In this equation, $L$ is the appropriately chosen Laplacian (or the Laplace-Beltrami operator on non-Euclidean domains).
%
\begin{figure*}[t]
\centering

\includegraphics[width=1\textwidth]{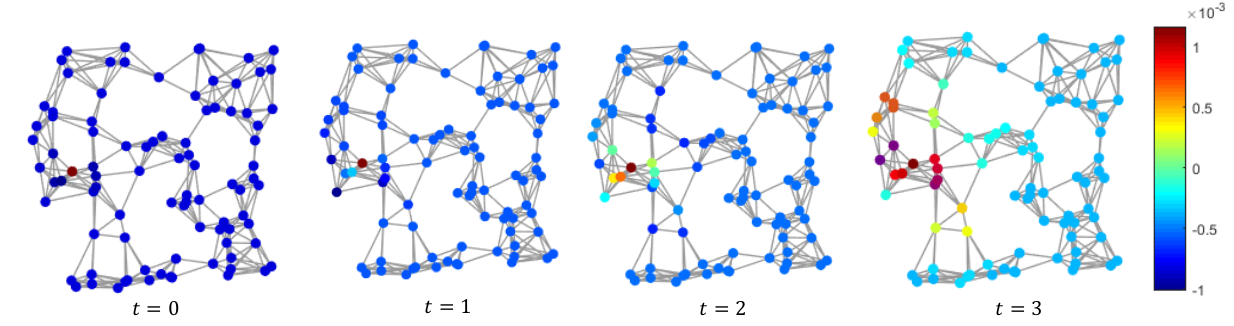}
\caption{Influence of different values of the time parameter $t$ in Eq. $~\eqref{eq:layer_tide}$ on the spatial extent of the output of time-derivative diffusion. Note how for larger time values, the spatial support increases.}
\label{fig1}
\end{figure*}
%
The solution to the diffusion equation is given by the heat operator $H_t$, so that $u_t = H_t(u_0) = \operatorname{exp}(-t L) u_0$. Importantly, the heat operator $H_t$ is differentiable with respect to $t$, which was recently used in \cite{sharp2022diffusionnet} to use the diffusion equation with \textit{a learnable time parameter} $t$ as a way to replace convolution and enable long-range communication in the context of learning on curved surfaces.

As anticipated earlier, our key idea is to also use the diffusion process for information propagation. However, instead of using the heat operator as in \cite{sharp2022diffusionnet} we exploit \textit{time-derivative diffusion} as a communication mechanism within graph neural networks. Taking the negative derivative of the heat operator with respect to $t$ we obtain:
\begin{align}
	\label{eq:time_deriv_diffusion}
	- \frac{\partial u_t }{\partial t} = L u_t = T_t(u_0) = L \exp(-t L) u_0,
\end{align}
\MKFinal{where $T_t (u)=  L H_t(u)$ is the \emph{time derivative diffusion} operator.}
We propose to use Eq.~\eqref{eq:time_deriv_diffusion} to diffuse information between the nodes. Specifically, we construct a single layer of our model that we call TIDE, within a graph neural network as follows:
\begin{align}
	\label{eq:layer_tide}
	\mathcal{L}^{\text{TIDE}}_k(U) = \MKFinal{\sigma \left( T_t(U)  W^{(k)} \right) =}\sigma \left( L \exp(-t_k L) U  W^{(k)} \right).
 \end{align}
Here $k$ is the layer index, $L$ is the Laplacian operator, and $W^{(k)}$ is the matrix of learnable weights associated with layer $k$.

Observe that our definition is similar to the standard message-passing layer defined in Eq.~\eqref{eq:layer_gcn}. However, crucially, our layer also includes the use of the diffusion operator $\exp(-t_k L)$ and in our resulting neural network architecture we make \textit{both}  the weight matrix $W^k$ and the layer-wise time parameter $t_k$  \textit{learnable parameters}.

Our layer, as defined in Eq.~\eqref{eq:layer_tide} has two major properties:
\begin{enumerate}
	\item First, by making the time $t_k$ a learnable parameter, we allow the network to optimize the spatial extent of the diffusion and thus enable potentially global communication across graph nodes.
	\item By using time-derivative diffusion instead of standard diffusion, we allow the network to revert to standard message-passing
 whenever necessary. Indeed, as shown
 \maxk{below}, our layer strictly generalizes the standard GCN layer, by simply setting $t=0$. Moreover, since we start neural network training by initializing all learnable parameters (including the learnable time in Eq.~\eqref{eq:layer_tide}) around zero, the resulting network can optimize the spatial extent of its output, \textit{only when necessary}.
\end{enumerate}

%
%
%
\mb{We utilize the augmented normalized adjacency matrix from  \cite{Kipf:2017tc} as the basis for diffusion in the following manner:}

%
%
\begin{align}
    \tilde{L} := \tilde{D}^{-\frac{1}{2}}\tilde{A}\tilde{D}^{-\frac{1}{2}}.
    \label{eq:laplace}
\end{align}

%
Here $\tilde{A} \in \mathbb{R}^{n \times n}$ is the adjacency matrix with self-loops (binary or weighted) and $D$ is the degree matrix with $\tilde{D}_{ii} = \sum_j \tilde{A}_{ij}$.

To provide intuition behind our approach, we illustrate in Fig.~\ref{fig1} 
 the spatial extent of the output of time-derivative diffusion, and thus of our layer $\mathcal{L}^{\text{TIDE}}$ defined in Eq.~\eqref{eq:layer_tide} depending on the time parameter $t$. Observe that for small values of $t$, the support is local and the output of $\mathcal{L}^{\text{TIDE}}$ is concentrated at the center node. However, for larger time values, the spatial support increases, thus facilitating distant communication between nodes. Crucially, as mentioned above, in our framework, we let the time parameter $t$ be a learnable variable, which allows the network to optimize it in a task-dependent manner.

\textbf{Remark 1:} \maxk{\textit{When the diffusion time $t_k$ is set to $0$, the diffusion block is equivalent to a GCN layer.}}
This follows directly from the properties of the operator exponential. Indeed, when $t_k=0$, we know that $\operatorname{exp}(-t_k L)$ is simply the identity operator $\mathbf{I}$. Thus, at $t=0$ we have $\mathcal{L}^{\text{TIDE}}_k(U) = \sigma\left(L U  W^k \right) = \mathcal{L}_k(U)$ as defined in Eq.~\eqref{eq:layer_gcn}.
%
%
%
%
which is one GCN layer, since we use the graph Laplacian operator $L$ in both cases.
From this simple result, we can conclude that our network is at least as expressive as a GCN model. Moreover, the weights of a message-passing model, such as GCN, can be directly translated into the model based on Eq.~\eqref{eq:layer_tide}, by using the same per-layer weight matrices and setting  $t=0$ in the latter. This simple flexibility allows our approach to increase the spatial support of every layer, only when it is useful for the underlying task.
\begin{figure*}[h]
    \centering
    \includegraphics[width = 1\linewidth]{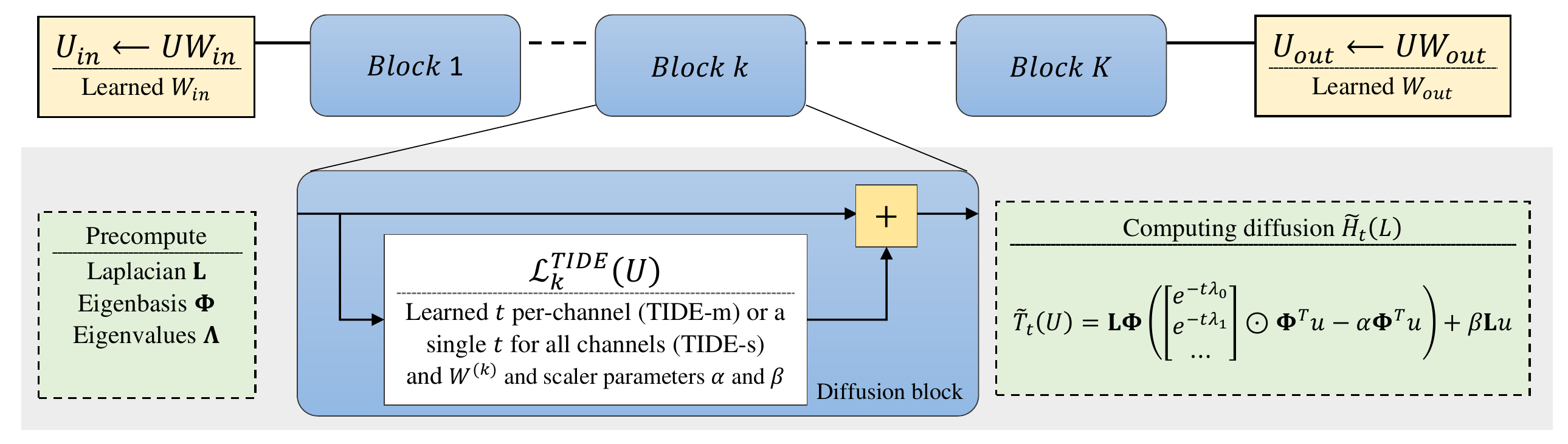}
    \caption{The framework of the proposed model. The network processes data from left to right. \maysam{Our architecture is composed of several time derivative diffusion blocks for information propagation and aggregation, as well as of the input/output layers to convert to the appropriate input and output dimensions.} 
    }
    \label{fig:architecture}
\end{figure*}

%
\paragraph{Computation of the diffusion}
As computing the matrix exponential of large graph Laplacian matrices is both computationally expensive and numerically unstable \cite{moler2003nineteen}, in our approach, we use spectral acceleration for the computation, as done in \cite{sharp2022diffusionnet}.
%
%
%
%
The key advantage of this formulation is that after a single pre-computation step, which calculates the Laplacian eigenbasis, the heat operator (and thus diffusion) for any time $t$ can be calculated by elementwise exponentiation.
%
%

%
The eigenvector problem of the Laplacian can be formulated as follows: $L \phi_i = \lambda_i \phi_i$, \MKFinal{where $\phi_i$ is the $i^{\text{th}} $ eigenvector of $L$ and $\lambda_i$ the corresponding eigenvalue sorted in ascending order by magnitude. }
After pre-computing the Laplacian eigenvectors and stacking \MKFinal{the first $l$ vectors} as columns of the matrix $\mathbf{\Phi}$, the heat operator can then be obtained as follows:
\begin{align}
	\label{eq:spectral_heat_op}
   H_t(u) =  \mathbf{\Phi} \begin{bmatrix}
    e^{-t\lambda_0} \\
    e^{-t \lambda_1} \\
    \dots
     \end{bmatrix}
     \odot \mathbf{\Phi}^\top u,
\end{align}
where $\mathbf{\Phi} = [\phi_i] \in
\mathbb{R}^{V \times l}$ and $\odot$
denotes the Hadamard product (elementwise
multiplication). In other words, we first
project the signal $u$ onto the basis
given by $\mathbf{\Phi}$ via $u
\rightarrow \mathbf{\Phi}^T u$. We then
multiply (in an element-wise manner) each
coefficient $i$ by $\exp(-t \lambda_i)$
where $\lambda_i$ is the eigenvalue
corresponding to the $i^{\text{th}}$
eigenvector and then convert back to the
standard basis by pre-multiplying by
$\mathbf{\Phi}$.

By this low-rank basis projection of the operator, some information is lost. To compensate for information loss, we introduce the operator $P = \mathbf{I} - \mathbf{\Phi} \mathbf{\Phi}^T$. For any signal $u$, note that $H_{0}(u) + P u = u$ holds. By incorporating the original signal mapped by $P$ into the spectral approximation of the heat operator $H_t(u)$, we can efficiently compensate for lost information. The modified heat operator is defined with two learnable scaling parameters $\alpha$ and $\beta$ as $\tilde{H}_t(u)= H_t(u)+(\beta  \mathbf{I} -\alpha \mathbf{\Phi} \mathbf{\Phi}^T)u$. After simplifying, the modified heat operator becomes:
%
%
%
%
\begin{align}
	\label{eq:spectral_heat_op2}
   \tilde{H}_t(u) =  \mathbf{\Phi}\left( \begin{bmatrix}
    e^{-t\lambda_0} \\
    e^{-t \lambda_1} \\
    \dots
     \end{bmatrix}
     \odot \mathbf{\Phi}^\top u - \alpha \mathbf{\Phi}^\top u   \right)+\beta u
\end{align}

Note that Eq.~\eqref{eq:spectral_heat_op2} is differentiable with respect to $t$, which is essential in our setting, as $t$ is learnable. 
\MKFinal{We thus define $\tilde{T}_t(u) := L\tilde{H}_t (u)$ in Eq.~\eqref{eq:layer_tide} and make both $\alpha$ and $\beta$ in  Eq.~\eqref{eq:spectral_heat_op2} learnable. }
\maxk{Note that this spectral approximation is only one possibility for computing time derivative diffusion and in Section \ref{sec:euler} we show that the diffusion equation can also be implemented with the Euler method.}

\subsection{Time Diffusion Analysis}
\label{sec:time_diff_ana}

The main contribution of our model is learnable time diffusion. In the following, we briefly overview the relation between diffusion time and the size of the neighborhood in the information propagation through the graph.
The diffusion Eq. \eqref{eq:heat_op} by the operator exponential $H_t = \exp(-t L)$ can be defined through its expansion as a Taylor series.
%
For $K$-hop diffusion, this series can be truncated to bound the diffusion to a $K$- hop neighborhood as follows:
\begin{align}
 u_t^K = \sum_{k=0}^{K} \frac{(-t)^k}{k !} L^ku
\end{align}
Therefore the signal gets diffused only within the $K$-hop neighborhood.
A simple bound on the difference between diffusion computed \textit{purely} by looking at the $K$-hop neighborhood and diffusion computed over the entire graph can be obtained as:
 \begin{align}
     \left\|u_t-u_t^K\right\|  
 \leq \sum_{k=K+1}^{\infty} \frac{|-t|^k C^k}{k !} \label{eq:approx}
 \end{align}
where $u$ is a normalized signal and $C$ is the biggest eigenvalue of $L$ (see also Appendix \ref{sec:threshold_for_diffusion}).

\textbf{Remark 2:} \textit{When considering the minimization of $\left\|u_t-u_t^K\right\|$ it is possible to minimize this term by either decreasing $t$ or increasing $K$}.

Therefore, intuitively, one can think of the diffusion time $t$ to correspond to the size of the neighborhood over which information is propagated. The smaller the time $t$, the smaller the neighborhood that can be used to to locally approximate diffusion over time $t$. Since in our architecture $t$ is learned by the network, we interpret it as enabling an \emph{optimizable receptive field} that can be adapted for different tasks and channels of the graph neural networks.

\subsection{Architecture}
In Fig.~\ref{fig:architecture} we show a schematic view of the proposed TIDE network architecture.
\maxk{Our network architecture is composed of three main parts: one \textit{Input layer}, one or more \textit{Time Derivative Diffusion block(s)}, and one \textit{Output layer}. A residual connection is adopted in each diffusion block to increase the accuracy and training performance, as shown in Appendix~\ref{sec:skip}.}
%
%
The network first processes each node feature individually with a linear layer.
Subsequently, the diffusion unit processes the graph features.
In each diffusion block, the network diffuses the node features of fixed $D$ channels. In the TIDE-m model, each channel has its own learnable time parameter while in the TIDE-s model a learnable parameter $t$ is common for all $D$ channels. Finally, the output linear layer converts the learned output to the expected output dimensions.

%
%

%
%

%% file: sec-experiments.tex
\begin{table*}[h]
\centering
\caption{Comparison of the accuracy of proposed models on several benchmarks with baseline methods (mean±std).}
\label{tab:table1}
\resizebox{\textwidth}{!}{%
\begin{tabular}{llllllll}
\hline
Model &
  \multicolumn{1}{c}{Cora} &
  \multicolumn{1}{c}{Citeseer} &
  \multicolumn{1}{c}{Pubmed} &
  \multicolumn{1}{c}{CoauthorCS} &
  \multicolumn{1}{c}{Computer} &
  \multicolumn{1}{c}{Photo} &
  \multicolumn{1}{c}{Ogbn-arxiv} \\ \hline
GCN \cite{Kipf:2017tc} & 83.30±0.36 & 68.23±0.91 & 76.78±0.31 & 90.17±0.50 & 81.01±0.65 & 91.71±0.67    & 65.91±0.12 \\
GAT \cite{velivckovic2017graph} & 81.83±0.42 & 69.19±0.53 & 75.49±0.43 & 90.15±0.35 & 80.25±0.52 & 91.57±0.41 & 54.23±0.22 \\
GRAND  \cite{chamberlain2021grand} & 80.71±0.86 & 68.06±0.18 & 74.61±0.25 & 90.59±0.21 & 72.96±0.49 & 84.17±0.34 & 59.29±0.12 \\
GCNII \cite{Ming2020}  &  79.94 ± 1.11  &  \underline{70.27±0.32}  &  76.59±0.7  &  84.27±0.80  &  32.63±8.6  &  57.41±3.6  &  49.87±0.37 \\
ACM \cite{luan2022revisiting} &  81.83±0.12  &  69.03±0.02  &  73.3±0.63  &  \textbf{91.50±0.13}  &  77±0.65  &  \textbf{92.42±0.29}  &  66.23±0.42 \\
DiffusionNet \cite{sharp2022diffusionnet} & 80.96±0.50 & 70.00±0.91 & 73.09±0.15 & 89.52±0.22   & 74.72±0.66 & 87.17±0.26 & 54.79±0.16 \\
TIDE-m & \textbf{84.47±0.43} & \textbf{70.32±0.68} & \textbf{77.59±0.04} & 89.86±0.30 & \underline{82.11±0.03}    & 91.33±0.47 & \underline{67.86±1.10}    \\
TIDE-s  & \underline{84.31±0.36}  & 70.24±0.80   & \underline{77.24±0.62}    & \underline{90.21±0.12} & \textbf{83.01±0.02} & \underline{92.06±0.51} & \textbf{68.43±0.35} \\ \hline
\end{tabular}
}
\end{table*}

\section{Experiments}
We compare our methods to strong baselines on typical node classification benchmarks and present novel long-distance communication experiments.

\paragraph{Setup} For a fair comparison, we set similar values for the common hyperparameters in all baselines. For this purpose, we use $64$ channels in the hidden layer and a dropout probability of $0.5$. All models are trained with a maximum of $500$ epochs with a learning rate of $0.01$.
\maxk{In our approach we choose the highest validation accuracy to determine the number of diffusion blocks used.}
%
More implementation details are described in Appendix \ref{sec:Implementation_details}.



%
%
%
%
\subsection{Node Classification}
\label{subsec:node_classification}
In the first experiment, we consider the standard node classification problem on a variety of benchmarks with different properties, described in Appendix  \ref{app-database},
and compare the performance of our proposed models, \maysam{TIDE-m and TIDE-s}, with several baselines. \maxk{For the experimental setup we follow the methodology of \cite{chamberlain2021grand}.}


Table \ref{tab:table1} reports the mean accuracy and standard deviation of 10 different runs of our proposed models equipped with baselines. \maysam{An ablation study for different numbers of blocks of the architecture is provided in Appendix \ref{app-abblation-blocks}.} The results obtained from most models are almost similar to their published numbers, except for GRAND \cite{chamberlain2021grand} which only achieves the reported numbers after fine-tuning the hyperparameters. In addition, we also include a baseline ``DiffusionNet'', which shares a similar design to our approach but uses the standard heat diffusion like in \cite{sharp2022diffusionnet} \maxk{in the diffusion block instead of our time-derivative diffusion.}

As can be seen in Table \ref{tab:table1}, the TIDE models outperform baselines in most benchmarks. 
This comparison demonstrates the ability of the TIDE model to take advantage of global communication. 
\mb{We note that while TIDE-m has more learnable time parameters making the network more flexible, these additional parameters can lead to overfitting, making TIDE-s more accurate in some scenarios.}
%



Most importantly, we observe a significant improvement compared to GCN \cite{Kipf:2017tc}, which forms the basis of our approach. \mb{This demonstrates the effectiveness of incorporating time derivative diffusion as a communication mechanism and suggests the potential of applying the TIDE method in combination with other graph neural network techniques in future research}. 
%
%



%
\subsection{\maxk{Long Range Communication}}
\begin{table*}[h]
\centering
\caption{Comparing different methods in the setting where feature vectors of unlabeled nodes are synthetically set to zero. }
\label{tab:table2}
\resizebox{\textwidth}{!}{%
\begin{tabular}{lcccccccccc}
\hline
Model & Cora & Citeseer & Pubmed & CoauthorCS & Computer & Photo  & Ogbn-arxiv & Average \\ \hline

GCN \cite{Kipf:2017tc}& 57.87 & 40.16 & 41.21 & 46.93 & \underline{59.82} & 74.09 & 56.65 & 53.82 \\
GAT \cite{velivckovic2017graph}& 56.19 & 40.73 & \textbf{45.26} & \textbf{50.59} & 53.14 & 66.74 & 42.47 & 50.73\\
GRAND  \cite{chamberlain2021grand}& 52.79 & 41.45 & 40.83 & 24.71 & 17.58 & 24.39 & 55.58 & 36.76 \\
DiffusionNet \cite{sharp2022diffusionnet}& 60.05 & \underline{60.08} & \underline{42.56} & 25.4 & 19.28 & 30.12 & 35.77 & 39.04 \\
TIDE-m &  \underline{76.85} & \textbf{61.45} & 41.98 & \underline{49.78} & 57.6 & \textbf{78.49} & \textbf{57.75} & \textbf{60.56} \\
TIDE-s & \textbf{77.46} & 60 & 41.95 & 40.90 & \textbf{60.21} & 77.4 & \underline{57.27} & \underline{59.31} \\

 \hline
\end{tabular}%
}
\end{table*}
As highlighted in Section \ref{sec:motivation}, message-passing GNNs suffer from oversmoothing, typically when using more than $2$-hop information propagation or more layers \cite{oono2019graph}.
In most cases, message-passing GNNs can apply two different layers for $2$-hop neighbors without oversmoothing.
Although this local information seems to be sufficient for some small-scaled citation graphs, for larger benchmarks long-distance communication can be useful.
To demonstrate the effectiveness of long-range communication, we \maysam{present two} experiments \maysam{on synthetic graphs}.

\maysam{In the first experiment, a synthetic graph for each benchmark is obtained by setting feature vectors of nodes that are not in the labeled training set to the zero vector}.
This scenario ensures that the final node labels are \textit{only calculated} by using the information computed via the message passed from the labeled nodes and not inferred by the linear parts of the model.
\MK{As seen in Table \ref{tab:dataset_stats}, Appendix \ref{app_long_range}, the average distance between unlabeled and labeled nodes is long enough to enforce long-distance communication}.


%

Table \ref{tab:table2} shows the results of this experiment.
This table further highlights the long-range communication capability of the proposed models, where TIDE models outperform the other GNNs in almost all datasets. Additional details are provided in Appendix \ref{app_long_range}.

\maysam{The second scenario is generated by the graphs introduced in \cite{Karimi2018}, which consists of 5000 nodes that are randomly partitioned into the train, validation, and test data with equal sizes. Inspired by the network with two groups and a homophily parameter $h$ in this paper, we generate $10$ different graphs by changing the $h$ in the range $0.0$ to $0.9$, with interval $0.1$.
The parameter $h$ indicates the likelihood of a node forming a connection to a neighbor with the same label.  Therefore, as $h$ increases, nodes prefer to connect with nodes of the same label, and thus with smaller $h$ the long-range dependency communications will be prominent.}
\begin{figure}[]
    \centering
\includegraphics[width = 0.96\linewidth]{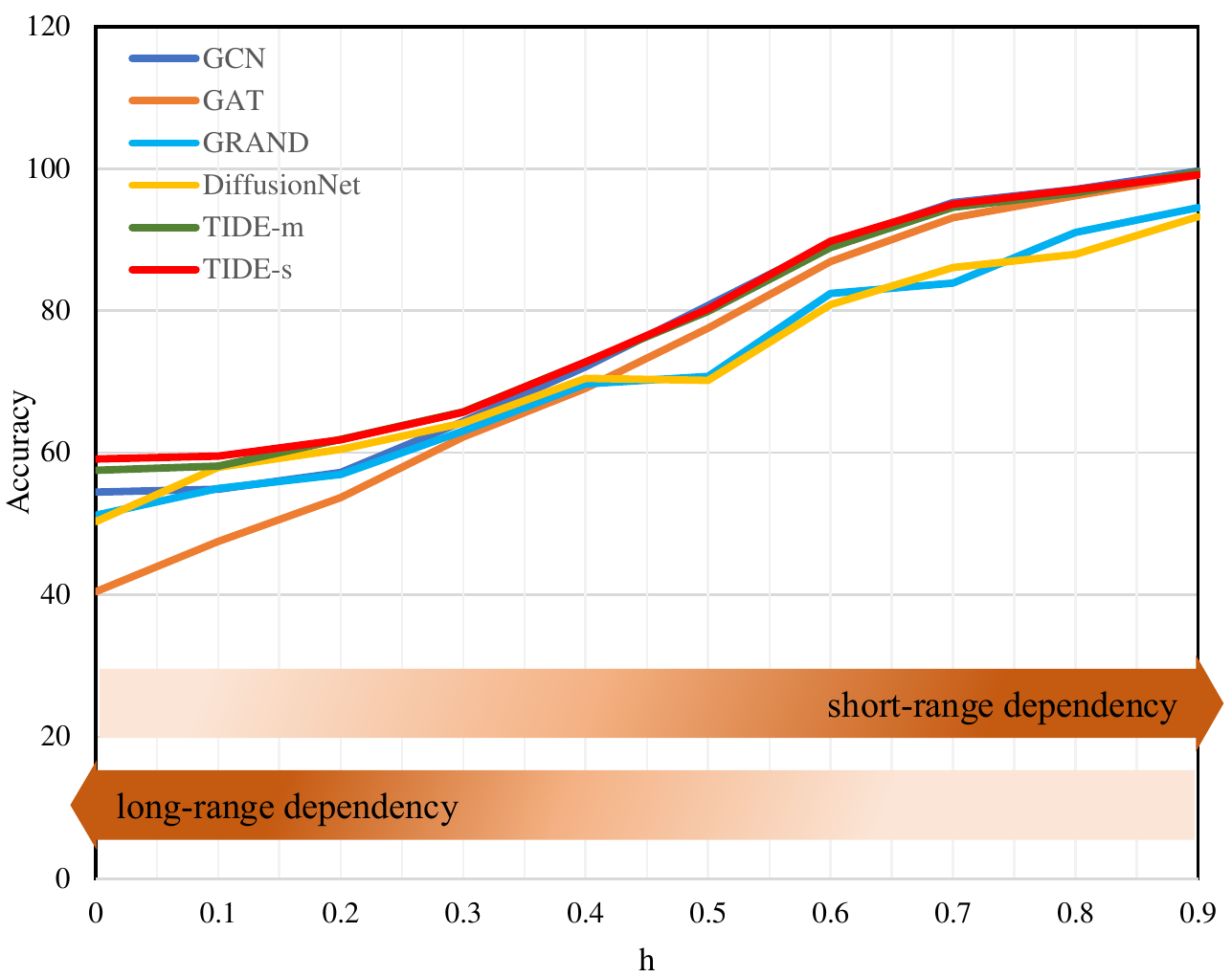}
    \caption{The accuracies of all baselines against the homophily parameter.}
    \label{fig:homophily}
\end{figure}
\maysam{The accuracies of all baselines against the homophily parameter are shown in Fig. \ref{fig:homophily}. As shown in the figure, with a smaller $h$, long-range communication between nodes is more necessary. 
The proposed TIDE-m and TIDE-s models perform significantly better than the baselines with local message passing. Note also that although all models perform better as $h$ increases, our approaches are the best-performing ones across \textit{every} value of $h$.}
\mb{The Appendix \ref{app_hetrophilic_graph} includes additional experiments conducted on graphs with different homophily rated. }


\subsection{Geometric Graphs}

One weakness of traditional graph neural networks is the capability of node classification on geometric graphs. 
%
%
The authors of \cite{bouritsas2022improving} mention that typically graphs neural networks have problems operating on regular graphs, such as grids, triangle meshes, etc. 
To evaluate the performance of our approach on such data, we build a synthetic dataset based on the FAUST \cite{Bogo:CVPR:2014} collection of shapes represented as triangle meshes.
We convert the meshes into graphs, by simply using the graph structure of the triangle mesh (i.e., using vertices of the mesh as graph nodes and edges of the faces as graph edges).
For the graph features, we will use the heat signature kernel (HKS) \cite{sun2009concise} with dimension $5$.
We develop three different synthetic settings: in \textbf{single} train and test nodes are on the same single graph,\maxk{ with a random $0.2$, $0.3$, $0.5$ split for train, validation, and test, respectively}. In the \textbf{multi} setting, we train the networks on the FAUST training set, which consists of $80$ graphs, and tested on the FAUST test set, consisting of 20 test shapes. In the \textbf{mixed} setting, we train on the FAUST training set and tested on SHREC'07 four legged setting.

\begin{figure}[t]
    \centering
\includegraphics[width = 0.96\linewidth]{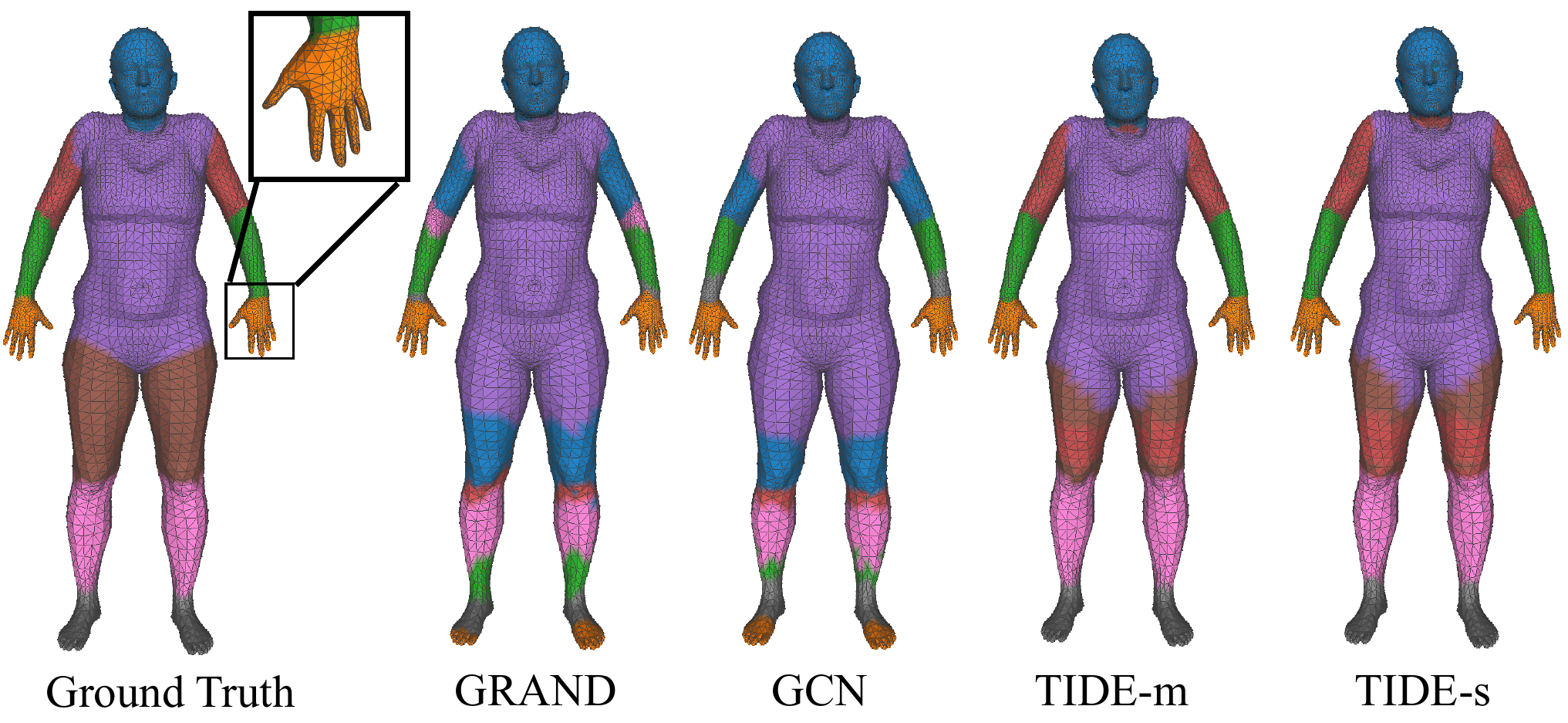}

    \caption{Qualitative example of labeling on one FAUST shape. \maxk{Note in particular how the other algorithms completely fail to label the upper leg and upper arms.}
    }
    \label{fig:FAUST_qualitative}
\end{figure}

We emphasize that for all baselines, we only use the graph structure of the shapes for learning and feature propagation, and ignore the node coordinates in 3D.
As we can see in Table~\ref{tab:FAUST} our approach outperforms the baselines GCN and GRAND by a  very significant margin in this scenario.
Note also that we evaluate the generalization capabilities of our method, by evaluating meshes retrieved from the entirely different four legged SHREC'07 \cite{giorgi2007shape} dataset.
%
%
Fig.~\ref{fig:FAUST_qualitative} shows a qualitative result in the  \textbf{single} scenario. We can see that for example, ours is the only method that is capable to recover the thigh as well as the upper arm.

\begin{table}[t]
 \centering
    \footnotesize
   \caption{ Testing the transfer capabilities of the graph networks on different datasets. \textbf{single}, \textbf{multi} and \textbf{mixed}.
    }
    \label{tab:FAUST}
     \begin{adjustbox}{max width=\linewidth}
    \begin{tabular}{lccc}
    \hline
    Model & single & multi & mixed\\
     \hline
    GCN \cite{Kipf:2017tc} & 69.21 &  65.90 & 65.83 \\
    GRAND \cite{chamberlain2021grand} & 78.46 & 69.54 & 46.32 \\
    TIDE-m & \textbf{94.11} & \underline{87.90} & \underline{81.22} \\
    TIDE-s & \underline{91.73} & \textbf{88.18} & \textbf{87.14} \\
    \hline
     \end{tabular}%
      \end{adjustbox}
\end{table}

\subsection{\mb{Computation Cost}}
\label{app_computational_cost}

\mb{The computation of TIDE can be divided into three distinct parts: preprocessing (\textit{pre}), training with derivative evaluation (\textit{train}), and inference evaluation (\textit{infer}). The preprocessing step includes the computation of Laplacian and eigendecomposition, which is executed only once on the CPU. For training and inference, TIDE utilizes standard linear algebra operations such as matrix multiplication and computation of the heat equation, which are efficiently executed on a GPU.} 

\mb{We conducted experiments to evaluate the runtime performance of TIDE and compared it with several baseline methods on graphs of varying sizes, including small, medium, and large-scale graphs. The results of these experiments are summarized in Table \ref{tab:table_Computational_cost}.}

\mb{According to the results presented in Table \ref{tab:table_Computational_cost}, the TIDE model performs comparably to spectral methods such as GCN and DiffusionNet in terms of runtime performance, while significantly outperforming the GRAND model. Nevertheless, the runtime performance achieved by the GAT models outperforms all the methods since it allows the model to selectively attend to relevant nodes in the graph and also shares parameters across all nodes. These results highlight the efficiency of TIDE, which achieves runtime performance comparable to other state-of-the-art methods.}

\begin{table}[t]
\caption{ Comparison of the runtime performance of TIDE and baseline methods on graphs of varying scales: Cora (2485 nodes), CoauthorCS (18333 nodes), and Ogbn-arxiv (169343 nodes). The reported runtime values are in seconds.}
\label{tab:table_Computational_cost}
\centering
\begin{adjustbox}{max width=\linewidth}
\begin{tabular}{llccc}
\hline
Model &  & \begin{tabular}[c]{@{}c@{}}Small \\ (Cora)\end{tabular} & \begin{tabular}[c]{@{}c@{}}Medium  \\ (CoauthorCS)\end{tabular} & \begin{tabular}[c]{@{}c@{}}Large \\   (Ogbn-arxiv)\end{tabular} \\ \hline
 & \textit{pre} & 0.5966 & 5.5038 & 213.79 \\
GCN & \textit{train} & 0.1217 & 0.1488 & 0.2411 \\
 & \textit{inference} & 0.0452 & 0.0736 & 0.107 \\ \hline
 & \textit{pre} & -- & -- & -- \\
GAT & \textit{train} & 0.0345 & 0.039 & 0.0921 \\
 & \textit{inference} & 0.024 & 0.0256 & 0.0414 \\ \hline
 & \textit{pre} & 0.5492 & 5.9851 & 209.19 \\
GRAND & \textit{train} & 1.5322 & 1.8333 & 2.9816 \\
 & \textit{inference} & 0.9646 & 0.5536 & 1.5923 \\ \hline
 & \textit{pre}* & 0.5650 & 5.5306 & 214.19 \\
DiffusionNet & \textit{train} & 0.0974 & 0.1131 & 0.1437 \\
 & \textit{inference} & 0.0375 & 0.0554 & 0.0679 \\ \hline
 & \textit{pre}* & 0.5650 & 5.5306 & 214.19 \\
TIDE-m & \textit{train} & 0.1188 & 0.1656 & 0.2551 \\
 & \textit{inference} & 0.049 & 0.0726 & 0.1099 \\ \hline
 & \textit{pre}* & 0.5650 & 5.5306 & 214.19 \\
TIDE-s & \textit{train} & 0.0975 & 0.1649 & 0.2344 \\
 & \textit{inference} & 0.0415 & 0.0747 & 0.0996 \\ \hline
\end{tabular}
\end{adjustbox}
* {\footnotesize Pre-processing times are the same for these approaches}.
\end{table}

\subsection{Ablation Studies}

\mb{We present ablation studies for Residual Connection and Diffusion Time in appendices \ref{sec:skip} and \ref{sec:diversity_of_t}, respectively.}

%% file: sec-conclusion.tex
\section{Conclusion, Limitations \& Future work}
In this work, we introduced a novel neural network architecture for graph learning.
Our key idea is to use time derivative with a learnable time parameter to augment the message-passing component of graph neural networks and enable long-range communication efficiently.
Our method is similar in efficiency to the strong GCN baseline, and scales well to large problem sizes, as no parameters depend on the number of nodes, and no expensive integration needs to be calculated during training.
As we build TIDE upon the standard message-passing paradigm, our approach is well-situated within the Weisfeiler-Lehman 1 category. As such, our current method cannot distinguish certain non-isomorphic graphs outside of this category. Nevertheless, we believe that the idea of time-derivative diffusion can be incorporated into other frameworks, such as recent methods with WL-3 expressive power. Moreover, it will also be interesting to extend our method to \textit{anisotropic diffusion} for information communication, while maintaining efficiency and differentiability of the time parameter. We leave this as an exciting direction for future work.


\section{Acknowledgements}
The authors acknowledge the anonymous reviewers for their valuable suggestions. Parts of this work were supported by the ERC Starting Grant No. 758800 (EXPROTEA) and the ANR AI Chair AIGRETTE.

%% file: sec-appendix.tex
\onecolumn
\appendix
\section{Appendix}

\subsection{Threshold for Diffusion}
\label{sec:threshold_for_diffusion}

The Taylor series for the diffusion equation is defined by:
\begin{align}
    u_t = e^{-tL}u=\sum_{k=0}^{\infty} \frac{1}{k !} (-tL)^ku = \sum_{k=0}^{\infty} \frac{-t^k}{k !} L^ku
\end{align}
Intuitively, for any $t \neq 0$, this series propagates information from the whole graph, as no factor in front of the power of the Laplacians is $0$.
\\
Eq. \eqref{eq:approx} can be derived simply as follows: 
 \begin{align}
     \left\|u_t-u_t^K\right\| & =\left\|\sum_{k=K+1}^{\infty} \frac{(-t L)^k u}{k !}\right\| \leq \sum_{k=K+1}^{\infty} \frac{|-t|^k C^k}{k !} \label{eq:approx_2}
 \end{align}

\subsection{Implementation Details}
\label{sec:Implementation_details}
\mb{The models are implemented in PyTorch, and the torch geometric library is also incorporated in addition to the standard PyTorch. To accelerate learning, GPU acceleration is utilized, while the diffusion operator and the gradient operator are preprocessed on a CPU using the SciPy library. The experiments are conducted on an NVIDIA A100 GPU with 40 GB of GPU memory. Despite the fact that the model has a small number of parameters, it can be trained on any GPU. 
}

\subsection{Datasets Properties}
\label{app-database}
The properties of different graph datasets used in the experiments are summarized in Table \ref{tab:table_datasets}.   

\begin{table*}[h]
\caption{Datasets properties}
\label{tab:table_datasets}
\centering
\resizebox{0.7\linewidth}{!}{%
\begin{tabular}{lccccccc}
\hline
Graph &
  \#Nodes &
  \#Edges &
  \begin{tabular}[c]{@{}c@{}}\#Node \\ featues\end{tabular} &
  \#Class &
  \begin{tabular}[c]{@{}c@{}}Avg.\\  node deg.\end{tabular} &
  \begin{tabular}[c]{@{}c@{}}Graph \\ diameter\end{tabular} &
  Label rate \\ \hline
Cora       & 2485   & 5069    & 1433 & 7  & 4.07   & 1.53 & 0.056 \\
Citeseer   & 2120   & 3679    & 3703 & 6  & 3.47   & 1.44 & 0.057 \\
PubMed     & 19717  & 44324   & 500  & 3  & 4.49   & 18   & 0.003 \\
CoauthorCS & 18333  & 81894   & 6805 & 15 & 8.93   & 24   & 0.016 \\
Computers  & 13381  & 245778  & 767  & 10 & 36.73  & 0.16 & 0.015 \\
Photos     & 7487   & 119043  & 745  & 8  & 31.79  & 0.23 & 0.021 \\
Ogbn-arxiv & 169343	& 1166243	& 128 & 40 &  13.67 &  23 &  1 \\
\hline
\end{tabular}
}
\end{table*}

\subsection{Number of Blocks and Oversmoothing }
\label{app-abblation-blocks}

\begin{figure}[h]
\centering
\includegraphics[width=0.9\linewidth]{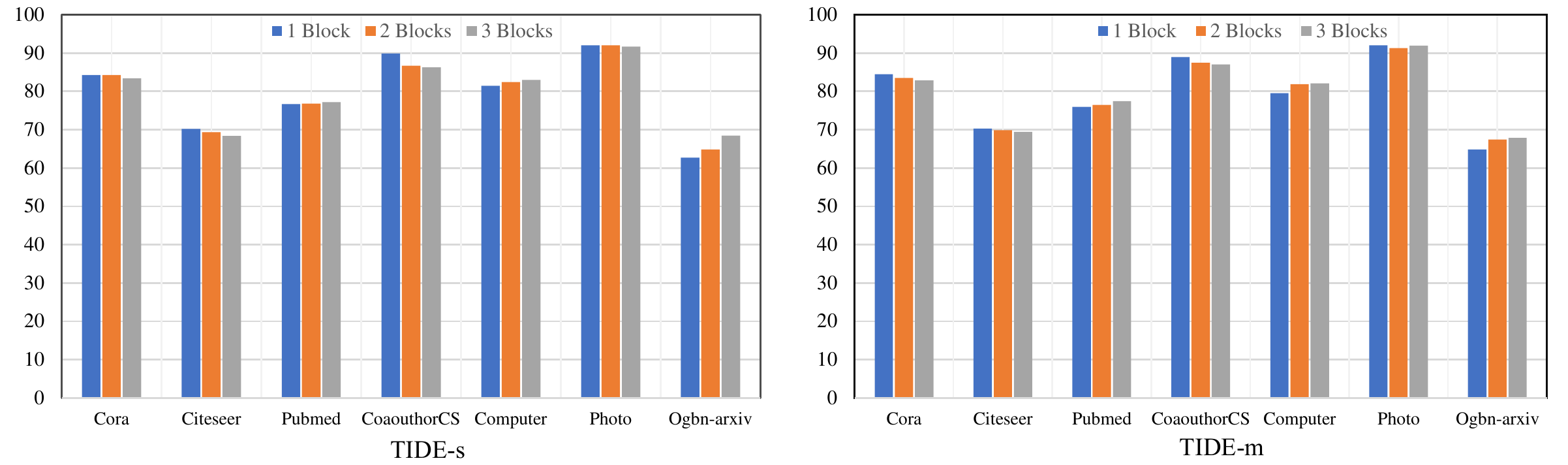}
\caption{Performances of the proposed models (TIDE-s and TIDE-m) by choosing the different number of blocks. }
\label{fig_blocks}
\vspace{-5pt}
\end{figure}

\mb{We assess the effectiveness of our proposed models by examining the performance across different numbers of blocks present in the architecture. Fig.~\ref{fig_blocks} illustrates that the network accuracy is influenced by the dataset when the number of blocks in the network is limited to 3 or fewer. Furthermore, we provide in Table~\ref{tab:oversmoothing} an analysis of how the test and validation accuracies vary as the number of diffusion blocks increases. The table illustrates that the TIDE-m model performs relatively well with up to 16 layers, and only begins to exhibit oversmoothing signs with 32 layers. We observe that our method maintains its high performance even with an increase in the number of diffusion blocks. Furthermore, our approach appears to be less prone to oversmoothing when compared to GCN.}

\begin{table}[ht]
\centering
\caption{\maxk{Performance of TIDE-m on the validation and test subset with $1,2,4, ... ,32$ diffusion blocks}  }
\label{tab:oversmoothing}
\begin{tabular}{clllll}
\hline
Dataset & \multicolumn{1}{c}{Model} & \multicolumn{1}{c}{$2^0$} & \multicolumn{1}{c}{$2^3$} & \multicolumn{1}{c}{$2^4$} & \multicolumn{1}{c}{$2^5$} \\ 
\hline
Cora & 
TIDE-m (\textit{test}) & \textbf{84.47±0.0} & 
\textbf{78.63±0.36} & \textbf{78.38±0.86} & \textbf{40.61±8.2} \\
 & TIDE-m (\textit{val}) & 82.49x0.41 & 78.74±1.1 & 78.08±1.1 & 40.55±9.2 \\
 & GCN (\textit{test}) & 72.57±0.23 & 74.03±0.42 & 57.11±0.78 & 27.78±0.44 \\
 & GCN (\textit{val}) & 74.51±0.39 & 75.04±0.19 & 55.32±0.24 & 32.99±0.89 \\ \hline
Citeseer & TIDE-m (\textit{test}) & \textbf{70.48±0.23} & \textbf{66.85±0.34} & \textbf{65.73±0.57} & \textbf{55.73±4.9} \\
 & TIDE-m (\textit{val}) & 71.97±0.2 & 67.30±0.36 & 66.47±0.0 & 52.20±5.2 \\
 & GCN (\textit{test}) & 61.05±0.43 & 57.23±0.34 & 56.55±0.23 & 22.43±0.81 \\
 & GCN (\textit{val}) & 63.23±0.81 & 58.01±32 & 58.62±0.92 & 19.25±0.54 \\ \hline
\end{tabular}
\end{table}

\subsection{Long-range Dependency}
\label{app_long_range}

To demonstrate the capability of our model in taking advantage of long-range communication, we present further insights into the scenario of zeroing out the test features. 
\maysam{To perform this experiment, we followed the data distribution used in the node classification experiment to determine the train, validation, and test data.}
In Table \ref{tab:dataset_stats}, we summarize the average and max distances between unlabeled nodes and the closest labeled one of each graph. According to these statistics in Table \ref{tab:dataset_stats}, the mean distance in each graph is quite far, which led to the fact that information aggregation is not possible by simply averaging from the neighbors. Therefore, long-range communication assumes paramount importance in this particular scenario. As shown in Table \ref{tab:dataset_stats}, the average distance between unlabeled nodes and labeled ones is more than $1$ on all datasets, and even on Citeseer, the average distance is $2.91$ which this value intelligibly reveals the reason for the significant improvement of TIDE compare with GCN in Table \ref{tab:table2}. \mb{Therefore, the $1$-hop neighborhood of nodes becomes ineffective since it yields a vector of zeros.}
The table demonstrates that for Photo, Computer, and Ogbn-arxiv the mean distance is smaller than a $2$-hop neighborhood. In these metrics, although the accuracy of TIDE is better than other baselines, we would expect a smaller accuracy increment, which is indeed the case. In addition, Table \ref{tab:table2} shows that with the bigger mean and max distance, TIDE outperforms previous methods. \mb{It is worth noting  that the limited percentage of labeled data in Pubmed (0.29\%) leads to a strong possibility of overfitting, and making conclusions difficult to interpret. As a consequence, we can conclude that the ability of the TIDE model to communicate over long-distances is evidenced by the $0$-node label experiment.}

\begin{table*}[h]
\centering
\caption{Statistics of graphs with  zeroing out the test feature. These statistics are the average and max distances between each unlabeled node and its closest labeled one on each graph. They reflect the long-range dependency among nodes in this scenario. }
\label{tab:dataset_stats}
\resizebox{0.4\textwidth}{!}{%
\begin{tabular}{lccc}
\hline
Graph &
Average distance & Max distance \\ \hline
Cora        & 2.39 & 9\\
Citeseer    & 2.91 & 13 \\
Pubmed      & 3.54 & 9\\
CoauthorCS & 2.41 & 7\\
Computer    &1.79 & 5 \\
Photo       &1.55 & 5\\
Ogbn-arxiv  &1.63 & 10 \\
\hline
\end{tabular}
}
\end{table*}

\subsection{\mb{Experiments on Graphs with Different Homophily Rates}}
\label{app_hetrophilic_graph}

\mb{In order to conduct a more comprehensive analysis, we investigate the performance of our proposed model on a series of heterophilic graphs, introduced in \cite{NEURIPS2021_ae816a80}.  These graphs exhibit varying levels of homophily and sizes, which are delineated in Table \ref{tab:table_homophily_datasets}. The homophily rate $h$ denotes the degree to which nodes in the graph connect with similar nodes (homophily) versus nodes with dissimilar nodes (heterophily).}

\mb{Table \ref{tab:table_homophily_performance} indicates that the proposed model achieved the best accuracy in six out of eight graphs. By comparing the results of the proposed model with the baselines, we can conclude that the proposed model outperforms the baselines on heterophilic graphs. Specifically, for larger graphs such as Snap-patents with a higher heterophily ratio, the proposed models achieved significantly higher performance scores than the baselines. This suggests that the proposed model is more effective in capturing the long dependency in heterophilic graphs, which is an important finding that can inform the development of more effective models for real-world scenarios.}

\begin{table*}[h]
\caption{Dataset properties. The parameter $h[0,1]$, represents the edge homophily ratio. When $h \rightarrow 1$  the graph exhibits high levels of homophily, whereas $h\rightarrow 0$, the graph displays strong levels of heterophily. }
\label{tab:table_homophily_datasets}
\centering
\resizebox{0.7\linewidth}{!}{%
\begin{tabular}{lcccccc}
\hline
Graph &
  \#Nodes &
  \#Edges &
  \begin{tabular}[c]{@{}c@{}}\#Node \\ featues\end{tabular} &
  \#Class &
  Class type &
  $h$
  \\ \hline
Chameleon  &  2,277  &  36,101  &  2,325  &  5  &  Wiki pages  &  0.23 \\
Actor  &  7,600  &  29,926  &  931  &  5  &  Actors in movies  &  0.22 \\
Cornell  &  183  &  295  &  1,703  &  5  &  Web pages  &  0.3 \\
Texas  &  183  &  309  &  1,703  &  5  &  Web pages  &  0.11 \\
Wisconsin  &  251  &  499  &  1,703  &  5  &  Web pages  &  0.21 \\
Genius  &  421,961  &  984,979  &  12  &  2  &  marked act.  &  0.618 \\
Twitch-gamers  &  168,114  &  6,797,557  &  7  &  2  &  mature content  &  0.545 \\
Snap-patents  &  2,923,922  &  13,975,788  &  269  &  5  &  time granted  &  0.073 \\
\hline
\end{tabular}
}
\end{table*}

\begin{table*}[h]
\caption{Comparison of the accuracy of proposed models on heterophilic graphs with baseline methods (mean±std). }
\label{tab:table_homophily_performance}
\centering
\resizebox{1\linewidth}{!}{%
\begin{tabular}{lcccccccc}
\hline
Model &
 Chameleon &
 Actor &
 Cornell &
 Texas &
 Wisconsin &
 Genius &
 Twitch-gamer &
 Snap-patents
  \\ \hline
GCN  &  45.18±0.62  &  29.38±0.5  &  43.24±1.3  &  63.51±1.9  &  54.92±9.7  &  80.87±0.13  &  \underline{60.60±0.19}  &  36.84±0.37 \\
GAT  &  44.96±6.2  &  28.88±1.0  &  54.05±1.1  &  62.16±0.08  &  55.88±1.4  &  79.83±0.23  &  53.08±0.16  &  38.76±0.75 \\
GRAND  &  50.33±0.47  &  35.00±0.28  &  55.41±1.9  &  \textbf{67.62±1.9}  &  64.86±1.3  &  82.47±0.08  &  59.85±0.03  &  38.89±0.42 \\
DiffusionNet  &  \textbf{53.84±1.1}  &  34.44±0.33  &  56.76±0.6  &  62.16±0.0  &  62.78±2.8  &  82.59±0.12  &  55.72±1.6  &  30.69±0.014 \\
TIDE-m  &  \underline{52.08±1.1}  &  \underline{36.18±0.47}  &  \underline{58.11±1.9}  &  \underline{64.86±1.5}  &  \textbf{69.61±1.4}  &  \textbf{83.03±0.06}  &  \textbf{60.81±0.04}  &  \underline{40.56±1.7} \\
TIDE-s  &  51.75±0.47  &  \textbf{36.64±0.47}  &  \textbf{59.46±1.9}  &  63.81±1.9  &  \underline{68.63±1.4}  &  \underline{83.01±0.06}  &  60.40±0.13  &  \textbf{40.75±0.58} \\

\hline
\end{tabular}
}
\end{table*}

\subsection{Ablation about Residual Connection}
\label{sec:skip}

\mb{Table \ref{tab:table_skip} presents the results of the ablation study conducted on the residual connection. The findings indicate that the combination of residual connection and diffusion is the most effective network architecture. Additionally, it is noteworthy that utilizing solely the residual connection without the diffusion model is significantly less effective.}


\begin{table*}[t]
\centering
\caption{Comparison of the effect of the residual connection on several benchmarks (mean±std)}
\label{tab:table_skip}
\resizebox{\textwidth}{!}{%
\begin{tabular}{llllllll}
\hline
Model &
  \multicolumn{1}{c}{Cora} &
  \multicolumn{1}{c}{Citeseer} &
  \multicolumn{1}{c}{Pubmed} &
  \multicolumn{1}{c}{CoauthorCS} &
  \multicolumn{1}{c}{Computer} &
  \multicolumn{1}{c}{Photo} &
  \multicolumn{1}{c}{Ogbn-arxiv} \\ \hline
TIDE-m & \textbf{84.47±0.43} & \textbf{70.32±0.68} & \underline{77.59±0.04} & 87.07±2.10 & 82.11±0.03    & 91.33±0.47 & \underline{67.86±1.10}    \\
without residual & 82.44±0.57 & 68.79±0.57 & \textbf{77.76±0.33} & \underline{88.47±0.15} &82.01±1.2 & 91.76±0.11 & 58.80±9.50 \\
TIDE-s  & \underline{84.31±0.36}  & \underline{70.24±0.80}   & 77.24±0.62    & 86.29±5.90 & \textbf{83.01±0.02} & \textbf{92.06±0.51} & \textbf{68.43±0.35} \\
without residual & 83.71±0.79 &68.39±0.23 &77.37±0.76 &  \textbf{88.73±0.06} & \underline{82.46±0.26} & \underline{91.87±0.52} & 64.60±0.67 \\ \hline
without diffusion & 54.89±0.00 & 55.59±0.01 & 66.20±0.00 & 86.47±0.00 & 61.61±0.00 & 77.02±0.00 & 56.60±0.70\\
\hline 
\end{tabular}
}
\end{table*}

\subsection{Direct Implicit Timestep }
\label{sec:euler}

\paragraph{Matrix exponential approximations in diffusion}
We show that our method does not depend on the type of diffusion approximation used. In particular, we compare the spectral approximation advocated in Section \ref{subsec:tide_method} with using the \textit{implicit Euler} scheme to simulate diffusion \cite{sharp2022diffusionnet}.
%

%
Since the graph Laplacians can get fairly big, we only mention the results of the small datasets, Cora and Citeseer as shown in  Table~\ref{tab:matrix_approximations}.
Observe that we obtain worse results using implicit Euler diffusion approximation, compared to when using spectral approximation. 
In addition, the latter benefits from the operator $P$, which recovers some of the information.
The first approach is mathematically exact. Hence we did not introduce any signal recovery mechanism here. However, with the huge matrices and possibility of singularities, the method is numerically less stable and results in worse results.
In addition, the spectral approximation method is more scalable to large graphs, which is why we adopt it throughout our work.

%
Instead of using the spectral acceleration to compute the time derivative diffusion, we can also use the implicit Euler method as follows:
\begin{align}
    H_t(u):=L(I+t L)^{-1} u.
\end{align}
It requires solving a sparse linear equation system for each iteration during training and testing.
The implicit version makes this approach numerically stable compared to the direct Euler version.
In PyTorch, it is possible to solve those systems and back-propagate through them.
However, on graph datasets, the Laplacian matrix can be rather large, and it is only with CPU memory possible to solve this equation system.
Hence, its training speed is drastically reduced.

In addition, we also found that only using the low-frequency approximation of diffusion is beneficial to regularize network training.

\begin{table}[h]
    \vspace{-15pt}
 \centering
    \footnotesize
   \caption{Using Spectral and implicit dense methods to calculate the diffusion equation
    }
    \label{tab:table3}     
     \begin{adjustbox}{max width=0.6\textwidth}
    \begin{tabular}{lccc}
    \hline
    Method & Cora & Citeseer\\
     \hline
    Spectral & \textbf{84.47±0.43}&\textbf{70.32±0.68}\\
    implicit dense & 82.18±0.00 & 68.60±0.01 \\
    \hline
     \end{tabular}%
      \end{adjustbox}
    \label{tab:matrix_approximations}
\end{table}


\subsection{Diversity of Learned Time $t$ and Channel Analysis}
\label{sec:diversity_of_t}

\begin{figure}[ht]
\centering
\includegraphics[width=0.7\linewidth]{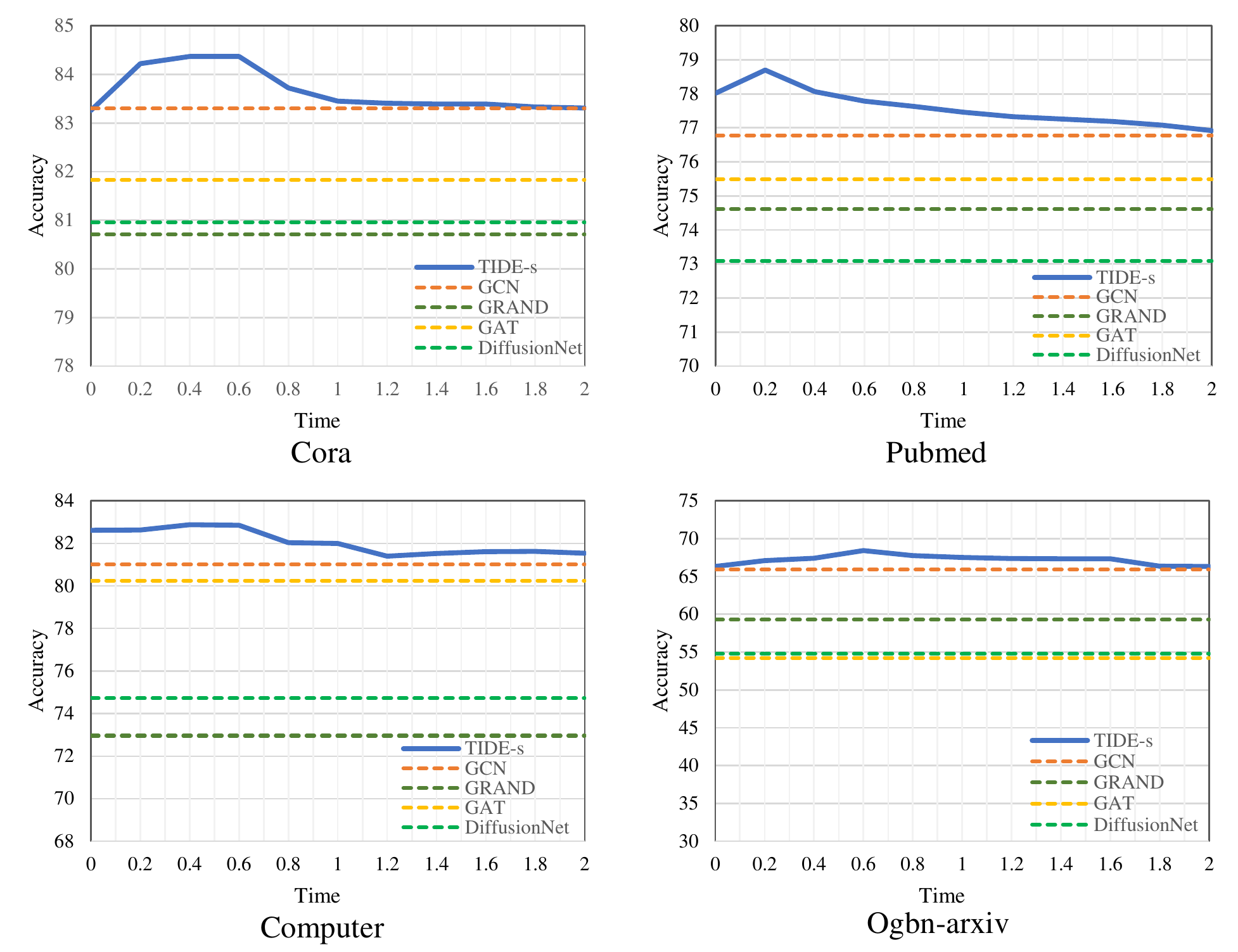}
\vspace{-10pt}
\caption{Comparison of accuracy of four benchmarks by changing the time parameter.}
\label{fig:time_ablation_2}
\vspace{15pt}
\end{figure}

The novel learning parameter of the TIDE model is time, which controls the diffusivity of the information of the node features. Since our model tries to find optimal times for each channel, to show how its performance can change depending on this parameter,  \mb{we explore a range of  $[0,2]$ with a step size of $0.1$ to identify the optimal times for each channel. In Fig.~\ref{fig:time_ablation_2}, we compare the accuracy of the TIDE-s model on four datasets at different fixed time values $t$. This experiment aims to evaluate how the choice of diffusion time affects the performance of the model and to determine the optimal value of $t$ for this particular dataset.}
We can see in this figure, for diffusion time $t=0$ the accuracy of our model and GCN are approximately equivalent. However, when $t$ is larger than zero, the best accuracy will be obtained with a specific value of time $t>0$ on each benchmark.

\begin{figure}[ht]
\centering
\includegraphics[width=\linewidth]{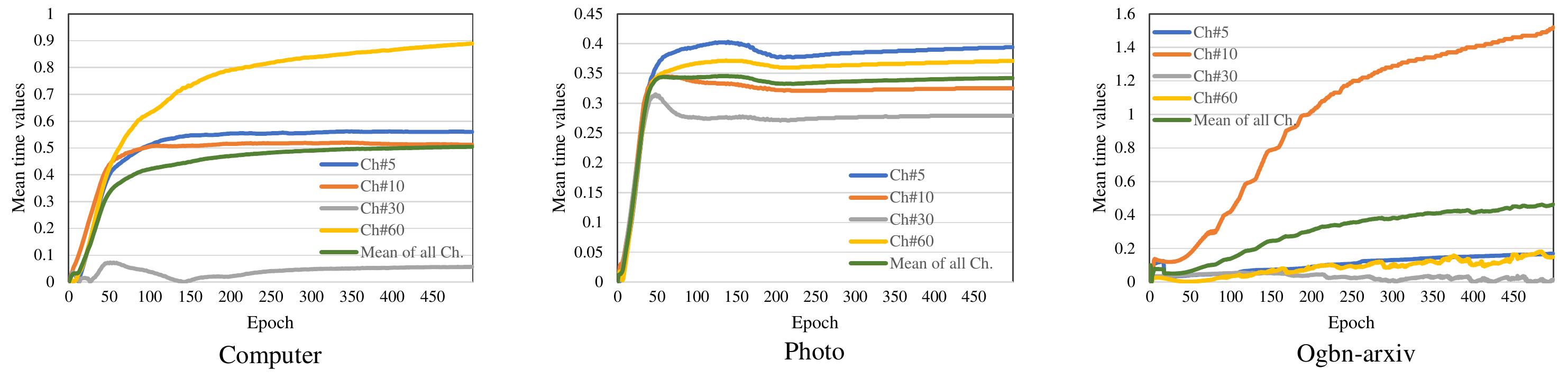}
\vspace{-20pt}
\caption{Value of time learned on channels 5, 10, 30, and 60 during the training. The green curve indicates the mean time values of these channels.}
\label{fig:channelGraph}
\end{figure}

\mb{In Fig.~\ref{fig:channelGraph} we can see the behavior of the learned time of the TIDE-m model on several sampled channels during the training on Computer, Photo, and Ogbn-arxiv datasets. It is observed that the learned time starts from a certain initial value and then gradually converges to an optimal value after an initial learning period.}

\textcolor{white}{.}